\documentclass{article}
\usepackage{spconf,amsmath,graphicx}

\usepackage{enumitem}
\setlist{nosep, leftmargin=14pt}
\usepackage{subcaption}
\usepackage{mwe} 
\usepackage{graphicx}
\usepackage{amsmath}
\usepackage{amsfonts}
\usepackage{multirow}
\usepackage{booktabs} 
\usepackage{bm}
\usepackage{url}
\usepackage{hyperref}
\usepackage{adjustbox}
\usepackage{threeparttable}
\usepackage[table,dvipsnames]{xcolor}
\usepackage{pifont}
\newcommand{\tablestyle}[2]{\setlength{\tabcolsep}{#1}
\renewcommand{\arraystretch}{#2}\centering\footnotesize}
\hypersetup{colorlinks=true, linkcolor=red, citecolor=blue}
\title{TAP-SLF: Parameter-Efficient Adaptation of Vision Foundation Models for Multi-Task Ultrasound Image Analysis}

\name{Hui Wan, Libin Lan$^{*}$}
\address{College of Computer Science and Engineering, Chongqing University of Technology, Chongqing, China\\
$^{*}$Corresponding author: lanlbn@cqut.edu.cn
}

\begin{document}

\maketitle

\begin{abstract}
Executing multiple tasks simultaneously in medical image analysis, including segmentation, classification, detection, and regression, often introduces significant challenges regarding model generalizability and the optimization of shared feature representations. While Vision Foundation Models (VFMs) provide powerful general representations, full fine-tuning on limited medical data is prone to overfitting and incurs high computational costs. Moreover, existing parameter-efficient fine-tuning approaches typically adopt task-agnostic adaptation protocols, overlooking both task-specific mechanisms and the varying sensitivity of model layers during fine-tuning. In this work, we propose Task-Aware Prompting and Selective Layer Fine-Tuning (TAP-SLF), a unified framework for multi-task ultrasound image analysis. TAP-SLF incorporates task-aware soft prompts to encode task-specific priors into the input token sequence and applies LoRA to selected specific top layers of the encoder. This strategy updates only a small fraction of the VFM parameters while keeping the pre-trained backbone frozen. By combining task-aware prompts with selective high-layer fine-tuning, TAP-SLF enables efficient VFM adaptation to diverse medical tasks within a shared backbone. 
Results on the FMC\_UIA 2026 Challenge \textit{test set}, where TAP-SLF wins fifth place, combined with evaluations on \textit{the officially released training dataset} using an 8:2 train-test split, demonstrate that task-aware prompting and selective layer tuning are effective strategies for efficient VFM adaptation.
Our code is publicly available at \href{https://github.com/huiwanHW/Florence-2-adaptation}{here}.
\end{abstract}


\section{Introduction}
\label{sec:intro}
Medical image analysis commonly necessitates the concurrent execution of multiple tasks, such as segmentation, classification, detection, and regression. Particularly within ultrasound clinical examinations, there is a growing emphasis on consolidating these tasks within a single unified framework to enable efficient multi-task processing.

While Multi-Task Learning (MTL) promotes shared representation learning across related tasks, medical imaging presents unique challenges due to the granular diversity of supervision signals, such as pixel-wise masks, image-level labels, bounding boxes, and scalar targets. In such scenarios, joint optimization is prone to gradient conflicts and negative transfer~\cite{Chen2018GradNorm, Yu2020PCGrad}. 
Recent efforts toward medical foundation models and large-scale pretraining further underscore the importance of unified representation learning~\cite{Azad2024FoundationSurvey, Azizi2024BiomedCLIP}. Thus, effectively disentangling shared representations from task-specific pathological signal remains a central challenge.

Vision Foundation Models (VFMs), such as Florence-2~\cite{Xiao2024Florence2}, provide robust pre-trained representations that generalize effectively across diverse domains. Nevertheless, full fine-tuning of these large models on limited medical data is computationally expensive and prone to overfitting. While Parameter-Efficient Fine-Tuning (PEFT) strategies, such as Low-Rank Adaptation (LoRA)~\cite{Hu2022LoRA} and Visual Prompt Tuning (VPT)~\cite{Jia2022Visual}, can significantly reduce trainable parameters and have demonstrated promising results in medical imaging contexts~\cite{Lian2024LessCouldBeBetter, HuangFPT_MICCAI2024}, 
their application in medical imaging remains nascent. Critically, most existing PEFT approaches employ task-agnostic, all-layer-uniform fine-tuning protocols, overlooking \iffalse the varying sensitivity of model layers \fi the layer-specific sensitivity to different medical tasks and supervision granularities. 

Given that different medical tasks emphasize distinct visual representations, such as segmentation relying on fine-grained spatial details, classification on global context, and detection on both localization and semantic discrimination, adaptation strategies should account for these diverse requirements. Additionally, prior transfer learning studies in medical imaging demonstrate that model layers exhibit varying contributions to downstream performance~\cite{Raghu2020Transfusion}. These observations suggest that effective VFM adaptation for multiple tasks should explicitly integrate both task-specific mechanisms and selective layer fine-tuning.

In this work, we propose \textbf{T}ask-\textbf{A}ware \textbf{P}rompting and \textbf{S}elective \textbf{L}ayer \textbf{F}ine-Tuning (\textbf{TAP-SLF}), a unified framework built upon the Florence-2 backbone~\cite{Xiao2024Florence2}. TAP-SLF introduces task-aware soft prompts to enhance task-specific performance and selectively injects LoRA modules into top encoder layers, updating 6.8\% of all parameters while keeping the bottom layers frozen. This strategy preserves the pretrained representations while enabling flexible, task-adaptive refinement.

The performance of TAP-SLF is validated on the official FMC\_UIA 2026 Challenge \textit{test set}, where our method ranks fifth place across four ultrasound image analysis tasks. Additionally, comprehensive evaluations on \textit{the officially released training dataset} using a standard 8:2 train-test split demonstrate performance gains across segmentation, classification, detection, and regression tasks, highlighting its robustness and generalizability.

% Our contributions are summarized as follows:
% \begin{itemize}
%     \item We propose TAP-SLF, a novel method for multi-task adaptation of vision foundation models. TAP-SLF enables parameter-efficient adaptation of VFMs to diverse medical imaging tasks within a unified shared backbone.
%     \item We propose task-aware soft prompting and selective layer tuning based on LoRA, a parameter-efficient fine-tuning strategy that enables vision foundation models, such as Florence-2, to be effectively adapted across multiple ultrasound image analysis tasks.
%     \item The fifth-place performance of TAP-SLF on the official FMC\_UIA 2026 Challenge \textit{test set}, alongside comprehensive comparisons and ablation studies on \textit{the officially released training dataset}, demonstrates its superior generalizability and effective adaptation across diverse ultrasound imaging tasks.
% \end{itemize}

\vspace{-1mm}

\section{Methodology}
\label{sec:method}
\subsection{Unified Network Architecture}
\begin{figure}[htbp]
\centering 
\includegraphics[width=1.0\linewidth, keepaspectratio]{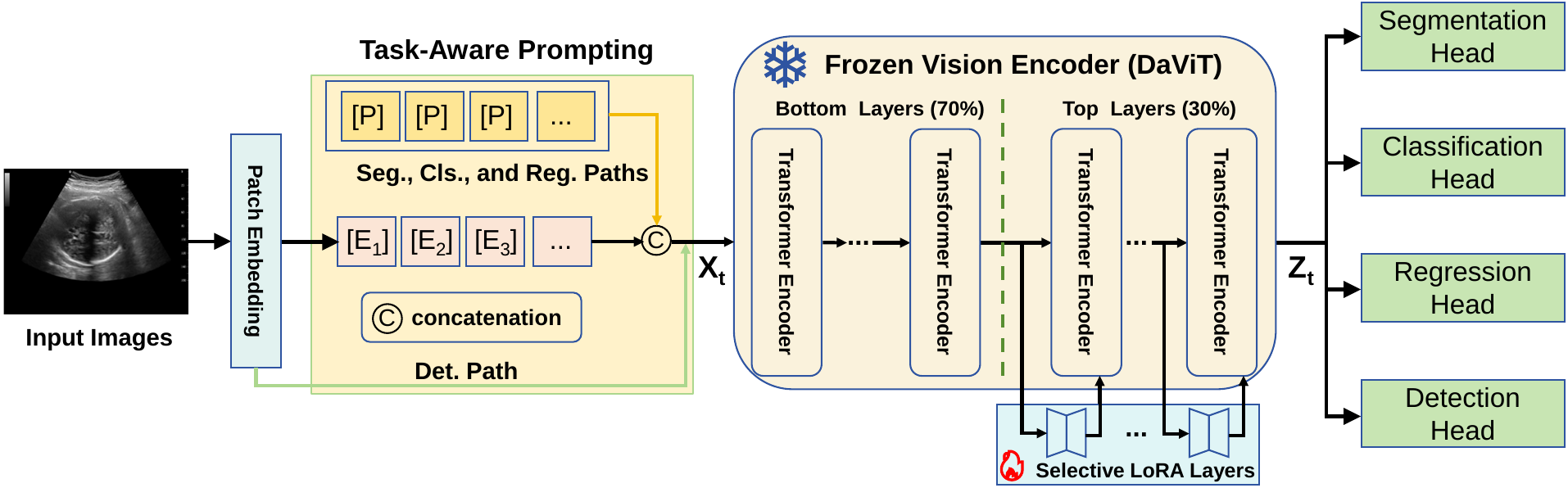}
\vspace{-3mm}
\caption{TAP-SLF \iffalse framework \fi overview. Input images are embedded into patch tokens. Task-aware prompts are prepended for segmentation, classification, and regression, while detection bypasses prompting to preserve spatial alignment. The transformer encoder uses selective \iffalse LoRA \fi layers fine-tuning, i.e., bottom 70\% frozen and top 30\% tuned. Task-specific representations $\mathbf{Z}_t$ are routed to \iffalse four\fi task-specific heads for final predictions.}
\label{fig:arch} 
\end{figure}

TAP-SLF is built upon the Florence-2 vision encoder~\cite{Xiao2024Florence2} and supports four tasks within a shared backbone: segmentation ($\mathcal{T}_{seg}$), classification ($\mathcal{T}_{cls}$), detection ($\mathcal{T}_{det}$), and regression ($\mathcal{T}_{reg}$). As shown in Fig.~\ref{fig:arch}, the framework comprises three key components: a task-aware soft prompting module that encodes task-specific priors at the input level, a selective layer tuning mechanism that injects LoRA adapters into top encoder layers, and four lightweight task-specific heads for output prediction. 
By leveraging a unified backbone, TAP-SLF enables highly parameter-efficient adaptation of the Florence-2 vision foundation model to multi-task ultrasound image analysis scenarios.
\vspace{-4mm}

\subsection{Task-Aware Soft Prompting}
To inject explicit task-specific priors into the shared vision encoder, we incorporate learnable soft prompts~\cite{Jia2022Visual}. These continuous, trainable vectors are prepended to the token sequence and optimized end-to-end alongside model parameters.

Formally, for each task $t \in \{\mathcal{T}_{seg}, \mathcal{T}_{cls}, \mathcal{T}_{reg}\}$, we define a set of $N$ learnable prompt vectors:
\begin{equation}
% \mathbf{P}_t = \{\mathbf{p}_{t,1}, \dots, \mathbf{p}_{t,N}\}, \quad \mathbf{p}_{t,i} \in \mathbb{R}^{D},
\mathbf{P}_t = \{\mathbf{p}_{t,1}, \dots, \mathbf{p}_{t,N}\}, \quad \mathbf{p}_{t,i} \in \mathbb{R}^{d},
\end{equation}
where $d$ denotes the model’s embedding dimension. The prompts $\mathbf{P}_t$ are concatenated with patch embeddings $\mathbf{E} \in \mathbb{R}^{L \times d}$ derived from input images via a patch embedding layer, yielding the initial token sequence $\mathbf{X}_0$:
\begin{equation}
\mathbf{X}_t = [\mathbf{P}_t ; \mathbf{E}] + \mathbf{Pos},
\end{equation}
where $[\cdot ; \cdot]$ denotes token concatenation along the sequence dimension, and $\mathbf{Pos} \in \mathbb{R}^{(N+L) \times d}$ represents the positional encoding that preserves the relative positions of tokens. 

During forward propagation, $\mathbf{X}_t$ undergoes self-attention across all transformer layers, enabling $\mathbf{P}_t$ to modulate attention toward task-discriminative features while keeping backbone weights frozen. This design achieves task-specific refinement without architectural changes.

Empirical observations reveal that soft prompting benefits $\mathcal{T}_{seg}$, $\mathcal{T}_{cls}$, and $\mathcal{T}_{reg}$ but degrades detection $\mathcal{T}_{det}$ performance. Since detection requires strict spatial correspondence for coordinate prediction, inserting prompt tokens would disrupt positional indexing. Therefore, we exclude prompts from the detection branch to preserve localization consistency.

\subsection{Selective Layer-Wise LoRA Injection}
To mitigate computational and memory constraints, TAP-SLF adopts selective layer fine-tuning via LoRA injection into a subset of transformer encoder layers, avoiding full backbone fine-tuning. LoRA approximates weight updates through trainable low-rank decomposition matrices injected into transformer layers. Formally, let $\mathbf{W}_0 \in \mathbb{R}^{d \times k}$ denote a pretrained weight matrix; LoRA parameterizes its update as:
\begin{equation}
\Delta \mathbf{W} = \mathbf{B}\mathbf{A},
\end{equation}
where $\mathbf{A} \in \mathbb{R}^{r \times k}$ and 
$\mathbf{B} \in \mathbb{R}^{d \times r}$ are trainable matrices with rank $r \ll \min(d,k)$.
% , $k$ is a bottleneck dimension

The updated projection weight is then formulated as:
\begin{equation}
\mathbf{W} = \mathbf{W}_0 + \Delta \mathbf{W}.
\end{equation}
Given inputs $\mathbf{X}_t$ to the linear projection layer, the LoRA-enhanced transformation produces task-specific representations $\mathbf{Z}_t$:
\begin{equation}
\mathbf{Z}_t = \mathbf{W}_0\mathbf{X}_t + \Delta \mathbf{W}\mathbf{X}_t=\mathbf{W}_0\mathbf{X}_t + \mathbf{B}\mathbf{A}\mathbf{X}_t.
\end{equation}

In TAP-SLF, LoRA modules are inserted into the attention projection layers, including query, key, value, and output projections, of the top 30\% of encoder layers, while the bottom 70\% of encoder layers remain frozen. This strategy preserves generic low-level features in the frozen bottom layers while enabling high-level semantic adaptation in the trainable top layers. This design benefits both dense prediction tasks, such as segmentation and detection, and global tasks like classification and regression, thereby enhancing overall task-specific performance.

This configuration ensures that only a minimal subset of parameters is trainable, substantially reducing memory footprint and computational cost relative to full fine-tuning.
\vspace{-1.5mm}

\subsection{Task-Specific Heads}
Upon obtaining task-specific representations $\mathbf{Z}_t$, TAP-SLF routes features through four task-specific prediction heads corresponding to segmentation, classification, regression, and detection. Following the baseline protocol~\cite{Deng2026Baseline}, global tasks like classification and regression employ a Global Average Pooling (GAP) module to aggregate features into task-level predictions. Dense prediction tasks such as segmentation and detection utilize a Feature Pyramid Network (FPN) decoder to aggregate multi-scale features and generate high-resolution feature maps.
\vspace{-2mm}

\section{Experiments and Results}
\label{sec:exp}
\subsection{Datasets}
\label{sec:datasets}
The FMC\_UIA 2026 Challenge \textit{test set}, which is held privately by the organizers and not publicly accessible, is used for official ranking evaluation. \textit{The officially released training dataset}, containing 40,387 images following an 80\%/20\% train-test split, is used for comparative evaluation of TAP-SLF and baselines. For all experiments on the released dataset, images are uniformly resized to 256$\times$256 before being fed into the network during training and inference.
\vspace{-2mm}

\subsection{Experimental Setup}
All experiments are conducted on a single NVIDIA GeForce RTX 3090 GPU with 24GB memory. Evaluation metrics include DSC and HD95 for segmentation, AUC/F1/MCC for classification, mIoU for detection, and MRE for regression. We adopt \texttt{florence-2}~\cite{Xiao2024Florence2} as the backbone. The prompt length $N$ is set to 10, and LoRA with rank $r=$ 8 and scaling factor $\alpha=$ 16 is applied to the top 30\% encoder blocks. Models are trained for 100 epochs with AdamW. Overall, TAP-SLF updates only 6.8\% of the total parameters.
\vspace{-2mm}

\subsection{Scoring and Ranking Result}
We submit our trained model to the organizers of the FMC\_UIA 2026 Challenge for evaluation on the private test set. The official results, summarized in Table~\ref{tab:rank}, show that TAP-SLF ranks 5th overall among all participating teams. Our method achieves top-tier performance in segmentation while maintaining competitive results across classification, detection, and regression tasks.
We observe that the performance gap relative to top-ranked methods primarily stems from the detection and regression branches. This suggests that further improvements in cross-task feature alignment and task interaction modeling could yield additional gains. Notably, TAP-SLF achieves this with only 6.8\% trainable parameters, balancing performance and efficiency.

\begin{table}[htbp]
\centering
\caption{Top-10 team rankings on the FMC\_UIA 2026 Challenge test set. Our team ranks 5th overall.}
\vspace{-2mm}
\label{tab:rank}
\resizebox{1.0\columnwidth}{!}{%
\begin{tabular}{l c c c c c c}
\toprule
Team & Seg. Avg & Cls. Avg & Det. Avg & Reg. Avg & Overall & Rank \\
\midrule
 UoN\_CVIP & 0.8931 & 0.9492 & 0.9621 & 0.9640 & 0.9421 &1 \\
 FDBME5008 & 0.8518 & 0.8891 & 0.9777 & 0.9942 & 0.9282& 2\\
 jinjingwu & 0.8794 & 0.8948 & 0.9630 & 0.9487 & 0.9215& 3\\
 FHLU & 0.9331 & 0.8072 & 0.9639 & 0.9686 & 0.9182 &4\\
 \rowcolor{SpringGreen}
 huiwan (Ours) & 0.9645 & 0.8241 & 0.9584 & 0.9155 & 0.9156 &5\\
Deeper & 0.8511 & 0.9718 & 0.8923 & 0.8992 & 0.9036 &6\\
 Lab105 & 0.8178 & 0.7912 & 0.9112 & 0.9128 & 0.8583 &7\\
 flamingo & 0.9819 & 0.8590 & 0.6690 & 0.9052 & 0.8538 &8\\
 ViCBiC & 0.7908 & 0.5060 & 1.0000 & 0.9949 & 0.8229 &9\\
Brain\_Team & 0.7122 & 0.8074 & 0.9198 & 0.8343 & 0.8184& 10\\
\bottomrule
\end{tabular}
}
\begin{tablenotes}
\footnotesize
\item [lef=0pt] FHLU: First Hospital of Lanzhou University
\end{tablenotes}
\end{table}
\vspace{-3mm}

\subsection{Comparison with Baselines}
We conduct comparative evaluations of TAP-SLF against established baselines, including the official challenge baseline~\cite{Deng2026Baseline}, Full LoRA~\cite{Hu2022LoRA}, and VPT~\cite{Jia2022Visual}. All methods are evaluated on the officially released dataset with an 80\%/20\% train-test split. Results summarized in Table~\ref{tab:results} demonstrate that TAP-SLF achieves consistent performance improvements across segmentation, detection, and regression tasks, while maintaining competitive performance on classification. Notably, this is accomplished while updating only 6.8\% of total parameters. These findings validate the effectiveness of our hybrid adaptation strategy for achieving robust multi-task performance under parameter efficiency constraints.

\begin{table}[htbp]
\centering
\caption{Performance comparison of different methods on the officially released dataset with \iffalse 80\%/20\% \fi 8:2 train-validation split.}
\vspace{-2mm}
\label{tab:results}
\resizebox{1.0\columnwidth}{!}{%
\begin{tabular}{lcccccccc}
\toprule
Method & DSC $\uparrow$ & HD95 $\downarrow$ & AUC $\uparrow$ & F1 $\uparrow$ & MCC $\uparrow$ & mIoU $\uparrow$ & MRE $\downarrow$ \\
\midrule
Official Baseline~\cite{Deng2026Baseline} 
& 0.7203 & 89.3289 & 0.8917 & 0.7950 & 0.6810 & 0.1632 & 60.1749 \\
Full LoRA~\cite{Hu2022LoRA} 
& 0.9211 & 10.3336 & 0.8983 & 0.7976 & 0.7185 & 0.6714 & 23.8229 \\
VPT~\cite{Jia2022Visual}
& 0.8064 & 59.8542 & 0.8512 & \textbf{0.8293} & \textbf{0.7628} & 0.3725 & 46.4173 \\
\textbf{TAP-SLF (Ours)} 
& \textbf{0.9423} & \textbf{7.2104} & \textbf{0.9038} & 0.8014 & 0.7012 & \textbf{0.6867} & \textbf{22.9845} \\
\bottomrule
\end{tabular}
}
\end{table}
\vspace{-6.5mm}

\subsection{Ablation Study}
\label{sec:ablation}
\subsubsection{Effectiveness of TAP and SLF}
We conduct an ablation study to evaluate the effectiveness of the TAP and SLF modules under identical training settings. The results are shown in Table~\ref{tab:component}. Using the pretrained backbone alone significantly degrades performance, confirming the necessity of encoder adaptation. Removing TAP primarily impacts segmentation and classification performance, while removing SLF adversely affects regression. Integrating both TAP and SLF achieves optimal performance across multiple metrics, particularly for segmentation. This demonstrates that our hybrid adaptation strategy can be effectively applied across diverse ultrasound image analysis tasks, as it preserves both low-level spatial details and high-level semantic information. Furthermore, results reveal that without TAP, MRE values decrease significantly while mIoU increases substantially, indicating that regression, as a global task, relies more heavily on deep-layer semantic adaptation, whereas detection demands precise spatial correspondence for accurate bounding box regression. Introducing prompt tokens would shift the positional indices of patch embeddings, potentially compromising spatial consistency. This finding directly motivates our task-aware prompting design, in which detection bypasses prompting to preserve spatial alignment.
\vspace{-1mm}

\begin{table}[htbp]
\centering
\caption{Ablation results of TAP and SLF under 8:2 train-test split.}
\vspace{-2mm}
\label{tab:component}
\resizebox{\columnwidth}{!}{%
\begin{tabular}{lccccccc}
\toprule
Configuration & DSC $\uparrow$ & HD95 $\downarrow$ & AUC $\uparrow$ & F1 $\uparrow$ & MCC $\uparrow$ & mIoU $\uparrow$ & MRE $\downarrow$ \\
\midrule
Backbone~\cite{Xiao2024Florence2} & 0.9072 & 12.1545 & 0.7410 & 0.7778 & 0.6797 & 0.4955 & 29.2690 \\
w/o TAP & 0.9402 & 7.4894 & 0.8921 & 0.7946 & 0.6916 & \textbf{0.6943} & \textbf{22.6911} \\
w/o SLF & 0.9408 & 7.3057 & 0.8901 & \textbf{0.8059} & 0.6947 & 0.6845 & 26.9228 \\
\textbf{TAP-SLF} & \textbf{0.9423} & \textbf{7.2104} & \textbf{0.9038} & 0.8014 & \textbf{0.7012} & 0.6867 & 22.9845 \\
\bottomrule
\end{tabular}
}
\end{table}
\vspace{-5mm}

\subsubsection{Effectiveness of Frozen Ratios}
We evaluate the impact of different frozen ratios on performance across all tasks. The results are summarized in Table~\ref{tab:sensitivity}. It can be seen that different tasks exhibit distinct sensitivities to the number of frozen layers. Segmentation and detection achieve optimal performance at a 70\% frozen ratio, indicating that preserving low-level representations in the frozen bottom layers yields superior spatial refinement. Classification and regression attain peak performance at a 50\% frozen ratio, suggesting a relatively high dependence on deep-layer adaptation. Excessive frozen, e.g., 80\% significantly degrades performance across all tasks. Overall, the 70\% frozen configuration achieves the most favorable performance-efficiency trade-off, supporting our design choice of selective layer-wise adaptation over uniform fine-tuning strategies.
\vspace{-1mm}

\begin{table}[htbp]
\centering
\caption{Ablation results of different frozen ratios under 8:2 train-test split.}
\vspace{-2mm}
\label{tab:sensitivity}
\resizebox{\columnwidth}{!}{%
\begin{tabular}{lccccccc}
\toprule
Frozen Ratios 
& DSC $\uparrow$ 
& HD95 $\downarrow$ 
& AUC $\uparrow$ 
& F1 $\uparrow$ 
& MCC $\uparrow$ 
& mIoU $\uparrow$ 
& MRE $\downarrow$ \\
\midrule
50\% 
& 0.9358 & 8.4126 & \textbf{0.9184} & \textbf{0.8221} & \textbf{0.7363} & 0.3612 & \textbf{21.4478} \\
60\% 
& 0.9286 & 10.2214 & 0.9082 & 0.8116 & 0.7195 & 0.5728 & 34.8821 \\
70\% 
& \textbf{0.9423} & \textbf{7.2104} & 0.9038 & 0.8014 & 0.7012 & \textbf{0.6867} & 22.9845 \\
80\% 
& 0.8814 & 28.7342 & 0.8675 & 0.7982 & 0.7094 & 0.6325 & 27.9913 \\
\bottomrule
\end{tabular}
}
\end{table}
\vspace{-6mm}

\subsection{Visualization}
\begin{figure}[htbp]
\tablestyle{0.45pt}{0.6}
   \begin{tabular}{ccc}
    Input&TAP-SLF&Ground Truth\\
    \includegraphics[width=0.33\linewidth]{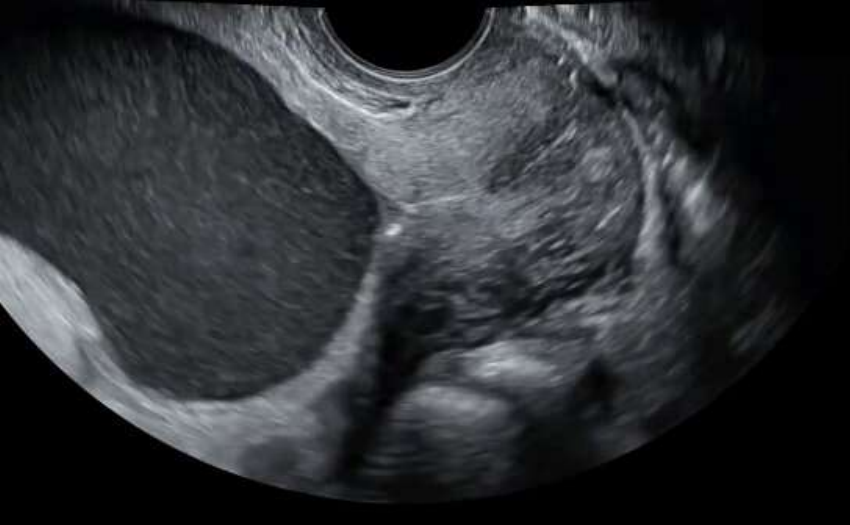}&
    \includegraphics[width=0.33\linewidth]{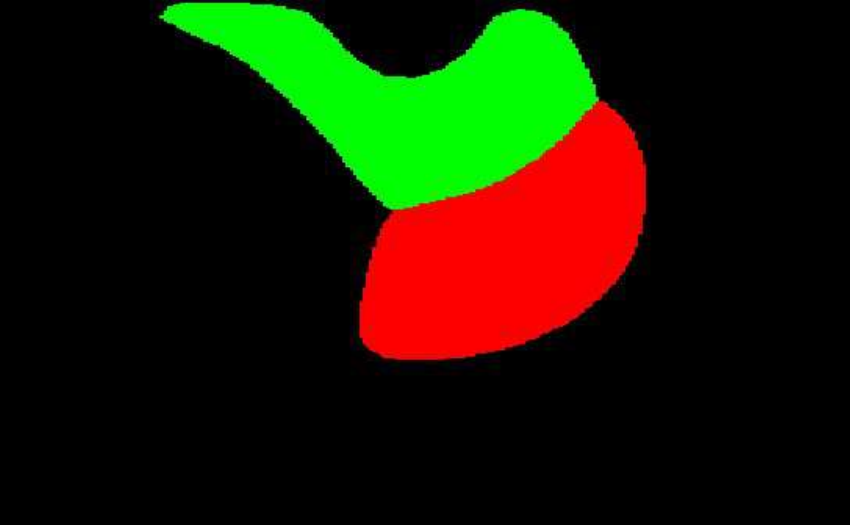}&
    \includegraphics[width=0.33\linewidth]{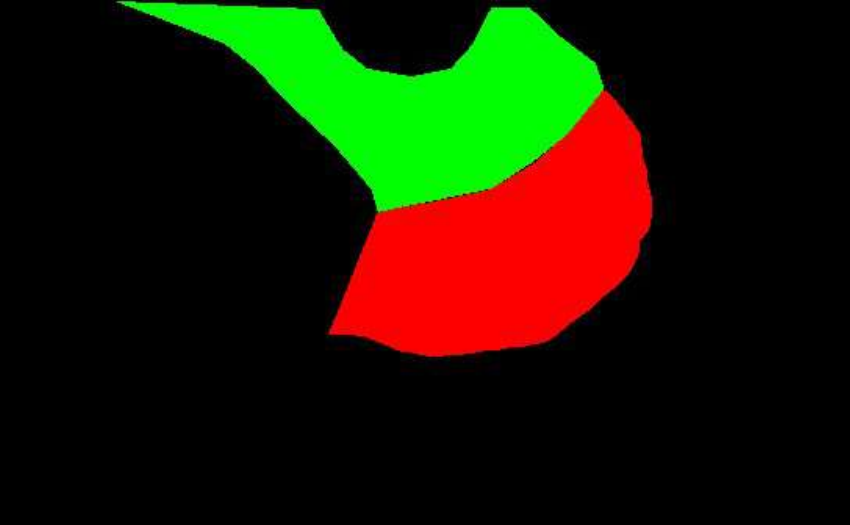}
    \end{tabular}\\[-0.5mm]
    \tablestyle{0.45pt}{0.6}
    \begin{tabular}{ccc}
    \includegraphics[width=0.33\linewidth]{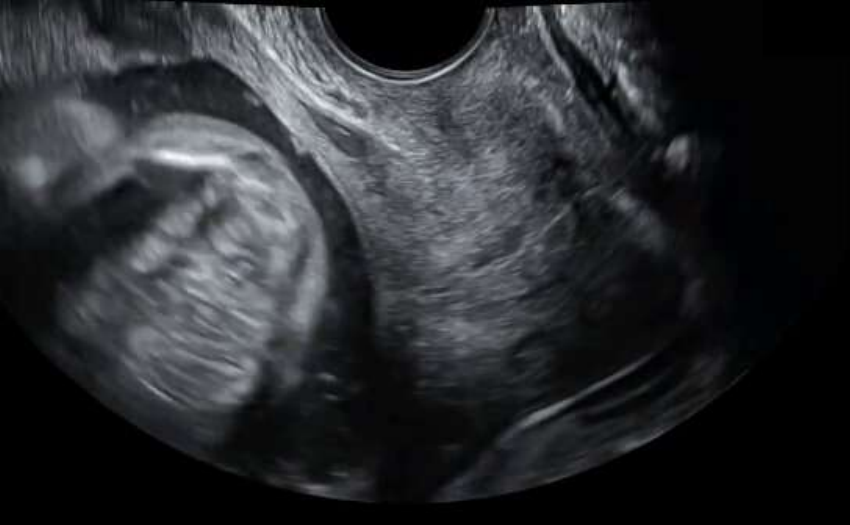}&
    \includegraphics[width=0.33\linewidth]{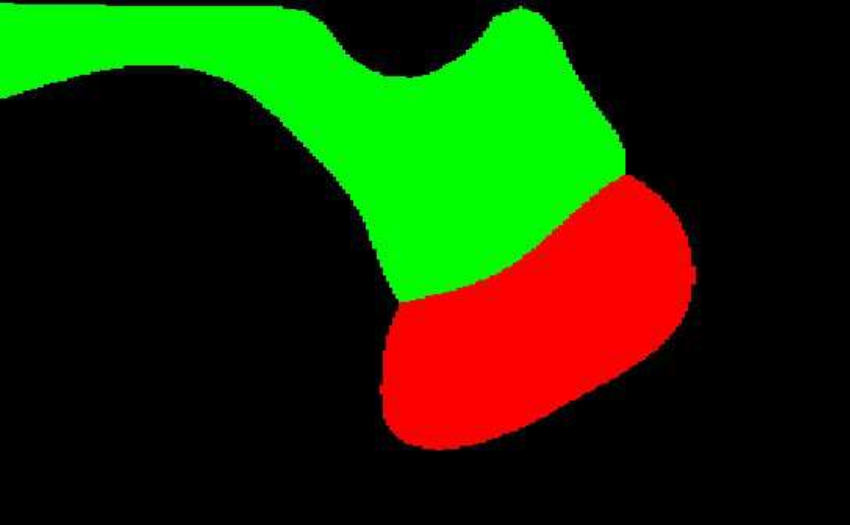}&
    \includegraphics[width=0.33\linewidth]{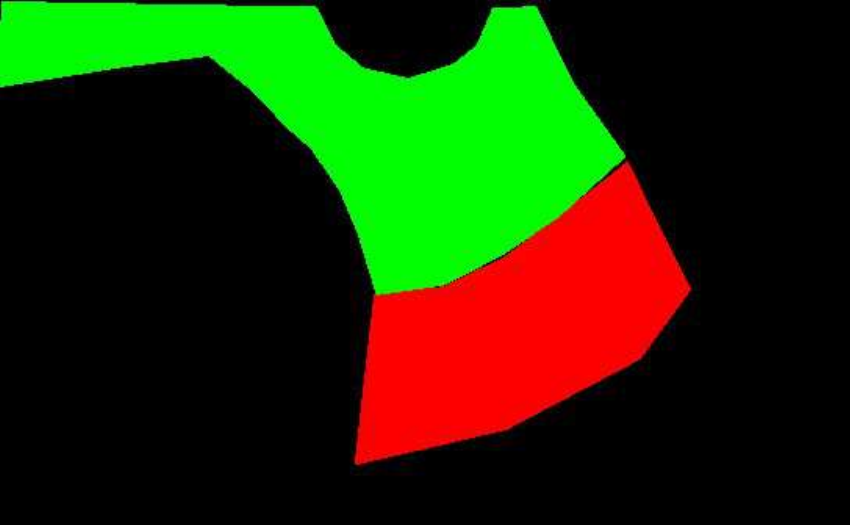}
    \end{tabular}\vspace{-2mm}
    \caption{The visualization results of TAP-SLF on a representative image from the maternal cervical segmentation dataset. The green and red regions indicate the segmented maternal anterior lip and posterior lip, respectively. Ground-truth annotations are shown for reference.} 
    \label{fig:main_results}
\end{figure}

Without loss of generality, we here present qualitative results for the segmentation task, compared against ground-truth annotations in Fig.~\ref{fig:main_results}. The predicted masks closely align with ground-truth annotations, accurately delineating target structures in ultrasound images.

\vspace{-1mm}

\section{Conclusion}
\label{sec:conclusion}
We introduce TAP-SLF, a parameter-efficient fine-tuning framework that integrates task-aware prompting with selective high-layer LoRA adaptation for multi-task ultrasound image analysis. By injecting task-specific prompts while preserving generic low-level representations and adapting task-sensitive higher layers, TAP-SLF wins fifth place on the FMC\_UIA 2026 Challenge and achieves nearly consistent improvements across segmentation, classification, detection, and regression on the officially released dataset. These results demonstrate that TAP-SLF provides an effective and efficient strategy for adapting vision foundation models to multi-task medical image analysis scenarios.

% References
% -------------------------------------------------------------------------
\bibliographystyle{IEEEbib}
\bibliography{ref}

\end{document}